\documentclass[runningheads]{llncs}
\usepackage[T1]{fontenc}
\usepackage{graphicx}
\usepackage{subcaption}
\usepackage{booktabs}
\usepackage{comment}
\usepackage{hyperref}
\usepackage[colorinlistoftodos]{todonotes}

\begin{document}
\title{Attribution and Uncertainty Behavior of Learned Residual Gyro Correction for Gyro-Stellar Estimation}
\titlerunning{Attribution and Uncertainty in Gyro Bias Correction}
% If the paper title is too long for the running head, you can set
% an abbreviated paper title here
%

\author{Mariela De Lucas Alvarez\inst{1}\orcidID{0000-0003-0846-4507} \and\\
Melvin Laux \inst{1} \orcidID{0000-0003-3517-7386}
\and
Arthur de Freitas Precht \inst{2}
\and
Maurice Martin \inst{2}
\and\\
Edoardo Caroselli \inst{2} \orcidID{0000-0002-4442-2219}
\and
Frank Kirchner \inst{1} \orcidID{0000-0002-1713-9784}
\and
Alexander Fabisch \inst{1} \orcidID{0000-0003-2824-7956}
}

\authorrunning{De Lucas Alvarez, M. et al.}

% First names are abbreviated in the running head.
% If there are more than two authors, 'et al.' is used.
%
\institute{Robotics Innovation Center, German Research Center for AI (DFKI GmbH), Bremen, Germany\\ \email{\{mariela.de\_lucas\_alvarez\}@dfki.de}\\
\and
Airbus Defence and Space GmbH, Immenstaad, Germany\\
% \email{\{abc,lncs\}@uni-heidelberg.de}
}

\maketitle              % typeset the header of the contribution
\begin{abstract}

This work investigates uncertainty decomposition and explainability in a deep learning-based framework for gyroscope bias correction. A 1-D Convolutional Neural Network is trained to predict residual angular rate corrections from multi-sensor inputs, including gyroscope and star tracker measurements. The bias corrections are sent to a flight-representative Gyro-Stellar Estimator. The network produces both mean corrections and input-dependent (heteroscedastic) aleatoric uncertainty, while epistemic uncertainty is estimated via an ensemble of independently trained models.

The proposed approach is trained under nominal conditions and evaluated in both nominal and structured perturbations that include additive and temporally correlated noise. Gradient-based attribution methods are applied to both the correction and uncertainty outputs, enabling a decomposition of the evidence that drives state updates and uncertainty estimates. By aggregating attribution patterns across rotational axes and regimes, we reveal axis-specific behaviors and characterize how structured perturbations influence the collaboration between aleatoric and epistemic uncertainty.

Uncertainty analysis shows that aleatoric uncertainty increases with perturbation intensity, but the distributions overlap and the calibration is not consistent across regimes. On the other hand, epistemic uncertainty gives a clear signal that gets clearer as the distributional shift happens, showing that the models disagree more. These results show that aleatoric and epistemic uncertainty work well together and that epistemic uncertainty is better at distinguishing between nominal and perturbed operating conditions. The results provide insight into the behavior of hybrid learning-based state estimation components and motivate the use of uncertainty for downstream monitoring and fault detection.

\end{abstract}
\section{Introduction}

Hybrid deep-filter architectures are continuously being adopted in spacecraft attitude estimation and control for improving accuracy, consistency, and performance \cite{Revach2022_kalmannet}.
These have demonstrated a clear contribution in robustness, particularly under partially known dynamics. However, there is still a lack of understanding of the behavior and the uncertainty of the learned components even under out-of-distribution conditions. This is crucial in safety-critical systems where interpretability is required, not optional.

Uncertainty quantification (UQ) can be used to address this limitation by decomposing uncertainty into aleatoric (input-dependent) and epistemic (model-dependent) components. 
However, calibration metrics such as coverage and likelihood primarily quantify predictive reliability and do not directly explain which input structures or model behaviors drive uncertainty estimates. Furthermore, most studies focus on independent and identically distributed (i.i.d.) or unstructured perturbations, missing important context when dealing with temporally correlated noise, which is characteristic of sensor degradation. Therefore, observed issues such as anisotropy and overlap between in-distribution (ID) and out-of-distribution (OOD) uncertainty still need to be explained.

At the same time, explainable AI (XAI) for timeseries has been widely used for input attribution in forecasting outputs. While these integrations are well established, their use in hybrid estimation pipelines remains limited. Additionally, it remains to be jointly used to explain uncertainty components together with learned corrections. These are the limitations that we address in this work.

We implement a Convolutional Neural Network (CNN) that learns residual corrections for gyro bias compensation. Gyroscope bias effects can persist even in the presence of classical filtering approaches, particularly under nonlinear operating conditions. Rather than replacing or modifying the internal estimation logic of the GSE, the CNN operates externally by predicting a residual gyro correction, enabling integration with existing state estimation pipelines while preserving the underlying estimator structure.

The predictor has an aleatoric output head that provides a mean correction $\mu$ and aleatoric uncertainty $\log\sigma^2$. Epistemic uncertainty is then calculated using an ensemble of independently trained models. The correction is then used in a Gyro-Stellar Estimator (GSE) and evaluated under nominal and structured perturbations such as additive and temporally correlated noise.

We combine attribution analysis with aleatoric and ensemble-based epistemic uncertainty evaluation to understand how correction and uncertainty behave and how different degrees of perturbations influence these factors. We use integrated gradients to evaluate the attribution of the aleatoric uncertainty and predicted mean under nominal and non-nominal conditions, allowing us to identify axis-specific sensitivities and the structured perturbations that interact between aleatoric and epistemic uncertainty across ID and OOD regimes.

\begin{figure}[tb]
    \centering
    \includegraphics[width=.99\linewidth]{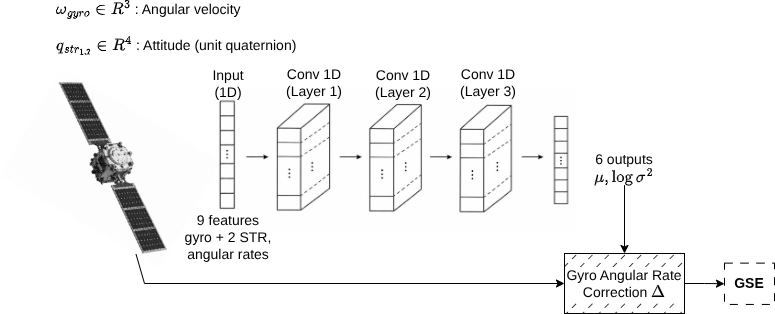}
    \caption{Overview of the analyzed hybrid estimation pipeline. A 1D-CNN processes angular-rate measurements from the gyroscope and two star trackers to predict a residual gyro correction term. The network outputs both the correction mean $\mu$ and aleatoric uncertainty $\log \sigma^2$. The predicted correction is subtracted from the gyro measurements before downstream state estimation with the GSE.}
    \label{fig:pipeline_summary}
    \vskip -0.4cm
\end{figure}

We present a systematic analysis of uncertainty and explainability behavior in a hybrid residual correction framework under structured perturbations. Our contributions are as follows:

\begin{itemize}
    \item Attribution of heteroscedastic output. We analyze the attribution in the contexts of sensor/time and $\mu$ vs $\log \sigma^2$ under perturbations. Prior work mostly focuses in deterministic outputs.
    \item Attribution of uncertainty. We analyze how uncertainty behaves with respect to calibration, entropy and epistemic separation to OOD response.
    \item Structured perturbation analysis. Most prior work focuses on i.i.d. noise, whereas we analyze multiple forms of temporally structured perturbations realistic to sensor degradation.
    \item Insight into uncertainty behavior. We provide a joint interpretation of aleatoric calibration and epistemic separation, highlighting attribution stability despite OOD and increasing uncertainty without feature redistribution.
\end{itemize}

Our findings highlight the influence of temporal correlations on uncertainty representation in spaceborne state estimation and offer comprehensible insights into the behavior of hybrid deep-filter systems beyond conventional calibration metrics.

\section{Related Work}

\begin{comment}
\subsection{Background}

Uncertainty decomposition:
\begin{itemize}
\item Kendall \& Gal \cite{Kendall_Gal_2017}: defines aleatoric (observation noise, irreducible) vs epistemic (model uncertainty, reducible with data); presents unified Bayesian DL framework combining both; NLL loss interpretable as learned attenuation; more robust to noisy data $\rightarrow$ framework we adopt
\item Nix \& Weigend \cite{Nix1994}: foundational heteroscedastic regression --- network jointly predicts mean and variance; derives NLL-based cost function for Gaussian error distribution $\rightarrow$ aleatoric head we implement
\item Lakshminarayanan et al.\ \cite{Lakshminarayanan2016}: deep ensembles as simple, scalable, well-calibrated epistemic estimator; shows higher uncertainty on OOD examples $\rightarrow$ ensemble approach we use for epistemic; relevant to ID/OOD regime analysis
\item Kuleshov et al.\ \cite{Kuleshov2018}: calibration procedure for regression (extends Platt scaling); guarantees calibrated credible intervals given enough data; tested on feedforward and recurrent NNs $\rightarrow$ calibration methodology we evaluate against
\end{itemize}

Attribution methods:
\begin{itemize}
\item Sundararajan et al.\ \cite{Sundararajan2017_ig}: Integrated Gradients --- satisfies Sensitivity \textit{and} Implementation Invariance axioms (most prior methods fail both); no network modification required; applies directly to CNNs $\rightarrow$ attribution method we use
\end{itemize}
\end{comment}

In this section, we review prior work on uncertainty-aware learning, explainability, and hybrid estimation methods relevant to learning gyro bias correction.

\subsubsection{Attribution methods}
Post-hoc attribution methods for neural networks include Layer-wise Relevance Propagation (LRP)~\cite{Bach2015_lrp}, which backpropagates relevance scores and targets image classifiers; Gradient-weighted Class Activation Mapping~\cite{selvaraju_grad-cam_2020}, which produces coarse spatial maps from gradients at the final convolutional layer; and Shapley Additive Explanations (SHAP)~\cite{lundberg_unified_2017}, a model-agnostic Shapley-value framework.

Although attribution methods have increasingly been adapted to time-series models~\cite{Mercier2022_benchmark,Jang2025_timing,Kechris2025_crossdomain}, comparatively fewer works analyze attribution behavior in sequential and uncertainty-aware time-series settings. Integrated Gradients (IG) \cite{Sundararajan2017_ig} is an attribution method that estimates an input feature's contribution to a model prediction. The method compares the prediction for a given input against a reference baseline (normally zeros or feature means) and accumulates the gradients along the interpolation path between baseline and input. They uniquely satisfy both Sensitivity (the full output difference relative to the baseline is attributed) and Implementation Invariance (functionally equivalent networks yield identical attributions).
Mercier et al.~\cite{Mercier2022_benchmark} benchmark attribution methods on time-series CNN classifiers and find that IG achieves a competitive performance among different gradient-based methods, particularly in terms of infidelity and runtime, while different attribution families exhibit trade-offs across evaluation criteria. Standard IG outperforms recent counterparts under metrics accounting for both positive and negative attributions~\cite{Jang2025_timing}.

\subsubsection{Learning-augmented state estimation.}
Learning-augmented state estimation improves accuracy under non-linear dynamics and partially known noise statistics by replacing or augmenting filter components with data-driven modules, e.g.,
KalmanNet replaces Kalman gain computation with an RNN that learns to operate with partial model knowledge while retaining the filter structure~\cite{Revach2022_kalmannet}, and
Al-Sharman et al.\ augment an extended Kalman filter (EKF) to identify and suppress measurement noise for attitude estimation~\cite{al-sharman_deep-learning-based_2020} of an unmanned aerial vehicle.
Our approach leaves the filter unchanged. We learn external residual bias corrections fed into a standard GSE, isolating the learned module and simplifying both deployment and analysis.
In the domain of spacecraft guidance, navigation, and control (GNC), prior work has applied SHAP-based XAI for adversarial attack detection~\cite{AI4GNC_City2024}. Our work shares this domain but focuses on uncertainty attribution under structured sensor perturbations representative of realistic degradation and operational variability.

\subsubsection{Uncertainty attribution}
Uncertainty can be split into aleatoric (noise of the data) and epistemic uncertainty (model uncertainty) \cite{Kendall_Gal_2017}.
Deep ensembles~\cite{Lakshminarayanan2016} provide well-calibrated epistemic uncertainty by training multiple independently initialized models, with reliably higher uncertainty on out-of-distribution inputs.
IG have been extended to uncertainty outputs where aleatoric and epistemic drivers are shown to differ systematically from those of mean predictions~\cite{Wang2023}. We build on this direction by analyzing attribution behavior in a hybrid deep-filter architecture under temporally structured perturbations, including axis-specific attribution shifts and uncertainty behavior across OOD regimes.

\section{Methodology}

In this section, we describe the methodology used to analyze uncertainty and explainability in learned gyro bias correction. We first introduce the CNN-based residual correction framework and the associated uncertainty representation used for probabilistic prediction. We then describe the structured perturbation modeling used to generate nominal and OOD operating regimes, followed by the attribution analysis pipeline employed to interpret both correction and uncertainty behavior.

\subsection{CNN-based Residual Framework}
The hybrid framework consists of a residual-learning CNN that feeds into the classical GSE. The core function of the CNN is to learn the residual bias correction using a multi-sensor time window as input comprised of the gyroscope and two calculated star tracker angular rates. The correction is learned externally to the GSE instead of changing the internal estimation logic directly. This design is motivated by observations from satellite test campaigns, where gyro bias effects persisted despite Kalman filtering, contributing to degradation in attitude knowledge under nonlinear operating conditions. The learning-based correction therefore aims to compensate these residual effects using available onboard sensor measurements while preserving the structure of the existing estimator. 

The CNN outputs are a residual mean correction vector and aleatoric uncertainty modeled as independent per-axis variances (see Figure~\ref{fig:pipeline_summary}). This hybrid formulation maintains the structure of the current estimator while allowing the neural network to correct gyroscope errors at the sensor level. Before state estimation, the predicted residual term is subtracted from the gyro measurements, preserving the existing pipeline.

The use of star tracker data in the input stream, together with gyroscope rates, provides complementary error characteristics to the network. The gyroscope provides high-frequency angular rate measurements but suffers from a low-frequency, time-varying bias drift that degrades attitude knowledge over time. Conversely, the star trackers provide drift-free, absolute attitude measurements but are subject to higher frequency noise and lower update rates. The stream units are homogenized by converting the attitude quaternions into angular rates.

In this manner, the model effectively gains a reference for \textit{ground truth} against which it compares the drifting gyroscope signal. This allows the network to isolate the bias component not just based on internal gyroscope dynamics (e.g., temperature or time correlations), but by observing the instantaneous discrepancy between the estimated rate from the star trackers and the biased rate from the gyroscope. 

The hybrid learning-filter architecture is set on an open-loop and evaluated at the residual level to provide a clear study without confounding effects from the estimator, providing a clear view of the bias correction behavior, uncertainty decomposition, and attribution patterns. 

\subsection{Uncertainty Representation}
We model heteroscedastic aleatoric uncertainty through a probabilistic output head and evaluate its calibration using standardized residuals and empirical coverage metrics. To capture predictive disagreement under distributional shift, we additionally employ ensemble modeling as an approximation of epistemic uncertainty. Finally, predictive entropy is used to characterize the sharpness and spread of the learned uncertainty distributions.

\subsubsection{Aleatoric uncertainty and calibration}
Aleatoric uncertainty is data-dependent and is integrated in the network enabling it to model input-dependent (heteroscedastic) noise directly. This approach follows the formulation introduced by Nix \& Weigend (1994) \cite{Nix1994} and later incorporated into deep learning by Kendall \& Gal (2017) \cite{Kendall_Gal_2017}, where the model learns not only the expected value of the regression target but also the uncertainty associated with that estimate. 

In this configuration, the network outputs now two quantities per target component:  the predicted mean correction $\mu(x)$ and the corresponding log-variance $s(x)=\log \sigma^2(x)$, capturing how noisy or ambiguous the input is. The network is trained using a Gaussian negative log-likelihood objective augmented with a weighted mean-squared error term,
\begin{equation}
    \mathcal{L} = \frac{1}{2} \left[ (y-\mu)^2 e^{-s} + s \right] + \alpha \, \mathrm{MSE}(y,\mu),
\end{equation}

where the first term corresponds to the heteroscedastic Gaussian likelihood and the additional weighted mean squared error (MSE) term stabilizes regression performance during training.

This loss encourages the model to assign higher variance to samples that are inherently difficult or ambiguous, while penalizing large residuals when variance is predicted too small. As a result, the network learns a realistic mapping from input features to observation noise, effectively capturing factors such as temperature-dependent behavior or calibration shifts despite these variables not being explicitly provided as inputs. This makes it an appropriate mechanism for embedding hidden or unobserved sources of variation, providing a proper way to adjust confidence based on the characteristics of each sample. 

The resulting output $(\mu, \sigma^{2})$  allows us to construct confidence intervals of the form $\mu \pm k \sigma$, which we use to analyze predictive reliability and calibration quality. Based on nominal coverage levels implied by the standard 68-95-99.7 rule of Normal distributions \cite{Kuleshov2018}: $\mu \pm 1\sigma$ should contain the truth $\approx68\%$ of the time, and $\mu \pm 2\sigma$ should contain the truth $\approx 95\%$ of the time.

\subsubsection{Ensemble Modeling}

Additionally, we perform ensemble modeling via independently trained ensembles to capture the model uncertainty. We train an ensemble of $M=5$ independently initialized models and with each prediction pair $(\mu_{m}, \sigma^{2}_m)$, we calculate epistemic uncertainty from the variance across predictions and compare ID/OOD distributions. We calculate the ensemble variance at each step for all ensemble members,
\begin{equation}
    u_{epi}(t) = \frac{1}{M} \sum^{M}_{m=1} (\mu_{t}^{(m)} - \bar\mu_{t})^{2},
\end{equation}
at each inference step $t$ and for each ensemble member $m$. The ensembles are used to ensure robustness of the subsequent attribution analysis. Instead of constructing a full predictive uncertainty, we analyze these components separately to understand their behavior under structured perturbations.

\subsubsection{Entropy as a Measure of Sharpness}
To summarize the overall sharpness of a predictive distribution, we compute the predictive entropy. While calibration provides an assessment of the predicted uncertainties, sharpness describes how concentrated the predictive distributions are regardless of ground truth. Sharper models produce narrower uncertainty bands (small $\sigma$), indicating confident predictions. Conversely, less sharp models produce broader distributions (large $\sigma$).

Because our aleatoric head outputs Gaussian variances, the predictive entropy is a direct monotonic function of $\sigma^{2}$.

For a Gaussian predictive distribution, the differential entropy~\cite{Cover2005_inftheory} provides a scalar summary of sharpness:
\(
    H = 0.5 \log (2 \pi e \sigma^{2}).
\)
Lower entropy corresponds to tighter, more concentrated predictions, while higher entropy reflects wider, more diffuse predictive distributions. We also use this metric during model selection.

Importantly, entropy only describes what the model believes about uncertainty, and does not indicate whether the uncertainty is correct. By comparing entropy and empirical coverage to nominal Gaussian levels, we evaluate whether the uncertainty estimates produced by the aleatoric head are well calibrated. 

In addition to evaluating $1\sigma$ and $2\sigma$ coverage levels and mean predictive entropy, we analyze standardized residuals ($z$-scores), defined as,
\(
    z = (y - \mu) / \sigma.
\)

Under ideal calibration, these residuals follow a standard normal distribution. Deviations from this behavior indicate model miscalibration, such as under- or over-dispersion in the predicted uncertainty.

\subsection{Perturbation Modeling}
We begin the perturbation generation with the nominal baseline. The ID data corresponds to nominal sensor conditions for both gyroscope and star trackers, without injected perturbations, reflecting standard operation conditions. For instance, the gyroscope exhibits approximately a $30^{\circ}$ drift consistently across all logs. These are all underlying dataset statistics.

The perturbation datasets are not used during training and are therefore out-of-distribution with respect to the training data.
We model these deviations to simulate sensor degradation and noise changes that could happen under real operation conditions. The perturbations are generated by modifying the statistical properties of the original underlying ID signals. In this manner, we preserve the underlying signal dynamics.

While these signals represent realistic sensor degradations, they do not exhaust the space of possible distribution shifts encountered in operation. Since the perturbations are not observed during training, the reported results should be interpreted as characterizing uncertainty behavior under controlled distribution shifts rather than demonstrating general OOD detection capability.

The perturbations are introduced in three intensity regimes: low, medium, and high, which progressively increase the noise or degradation severity, and are categorized as follows:

\subsubsection{Additive noise} This is temporally correlated and models stochastic sensor noise. Applied to both gyroscope and star trackers. Additionally, we include periodic vibrations that aim to mimic mechanical oscillations. We use a first order auto-regressive (AR(1)) process \cite{Box2016_timeseries,Bishop2006_PRM}
\(
    z_{t} = \phi z_{t-1} + \alpha_{t},
\)
where $\phi$ is the auto-regressive coefficient, and $\alpha_{t} \sim \mathcal{N}(0, \sigma^{2})$ is the innovation noise. To ensure a stationary variance of $\sigma^{2}$, the innovation variance is chosen as $Var(\alpha_{t}) = (1-\phi^{2})\sigma^{2}$.
Additionally, we model periodic vibration as a sum of sinusoidal components at predefined frequencies, $ft$, and random phase $\phi_f$ \cite{Lalanne1988_vibration},
\begin{equation}
    v(t) = \sum_{f \in \mathcal{F}} A \sin(2\pi f t + \phi_f),
\end{equation}
allowing us to inject oscillatory disturbances.

\subsubsection{Bias step and drift} These are slow-varying bias processes. This is applied only to gyroscope sensor readings. The dataset already contains a baseline drift of approximately $30^{\circ}$. The additional bias is meant to simulate unobserved effects such as sensor degradation or temperature variations. We consider a constant bias offset, $b_0$, and a drift rate per sample, $r$, and use
\(
    b_t = b_0 + rt.
\)

\subsubsection{Dropout} This only applies to star tracker sensors using a hold-last-value strategy. This is meant to realistically reflect intermittent attitude update freezes due to tracking loss, introducing piecewise temporal structure.

All types of perturbations are synthetic and structured, introducing temporal dependencies that reflect realistic sensor degradation processes. We use these to evaluate the response of uncertainty estimates and how meaningfully they react when the sensor data deviates from the training distribution. To do this, we retrain the ensembles with these perturbations while maintaining the ground truth bias unchanged. This ensures that only the uncertainty awareness is tested to capture epistemic uncertainty through prediction disagreement, rather than evaluating robustness in terms of accuracy. 

\subsection{Attribution Analysis}
Our explainability approach is driven by a mechanistic analysis of the CNN-bias outputs. We use gradient-based attribution to identify the main drivers of the mean residual correction and aleatoric uncertainty. These are analyzed under nominal (ID) and perturbed (OOD) conditions. This supports the understanding of the internal behavior of the model, not just performance.

We use integrated gradients to observe axis-specific behavior. The attributions are used to estimate the contribution of each input feature to the model's output. This is applied to the model outputs using time windows as input, enabling axis-wise timeseries analysis that captures sensor sensitivities.

Each attribution is computed with respect to each output. For each output dimension we consider the mean correction $\mu_i$, and the uncertainty $\log\sigma^{2}_i$, and we evaluate over $N$ input samples. The resulting attribution maps quantify the contribution of each input channel and timestep to the selected output. By doing this independently for each axis $i \in \{x, y, z\}$ we are able to decompose important sensor behaviors that influence each component of the aleatoric head.

We start with a raw attribution map per sample of size $N \times C$, where $N$ is the temporal window and $C$ is the number of input features (gyroscope and star trackers). To obtain interpretable quantities, the attributions are aggregated along each dimension. Channel importance is computed by summing over time, while temporal importance is obtained by summing over channels. This procedure reduces the dimensionality of the attributions while preserving their dominant structure.

The resulting attribution maps remain noisy at the per-sample level. To obtain more stable estimates, they are averaged across samples. This aggregation is performed separately for each axis and perturbation regime, ensuring that differences in model behavior are preserved across conditions. The resulting averages provide representative attribution patterns for each operating regime.

We compare attribution patterns across axes ($x$ vs $y$ vs $z$) and across operating regimes (ID vs OOD). This comparison is performed on both channel-wise importance and temporal attribution structure. By analyzing these dimensions, we assess how different inputs contribute to the predicted corrections and uncertainty estimates. Differences in attribution patterns across regimes reveal shifts in model behavior induced by structured perturbations.

The goal of this analysis is to identify the main drivers of the model's output. This enables characterization of the mechanisms underlying the predicted mean correction. Similarly, analysis of the uncertainty output allows comparison between mean and variance behavior, highlighting potential differences in their dependence on the input. By considering multiple regimes under structured perturbations, we are able to assess how these mechanisms shift across conditions. This allows us to highlight any correlations in the corrections and uncertainties.

\section{Experiments}

Our experiments evaluate both correction performance and uncertainty behavior under structured perturbations. While performance metrics quantify the robustness of the learned correction model under ID conditions, the primary objective is to analyze calibration, uncertainty separation, and attribution behavior as perturbation intensity increases.

We first describe the dataset and experimental setup, followed by an assessment of correction performance relative to raw measurements and a naive baseline. We then analyze aleatoric and epistemic uncertainty behavior across regimes before concluding with attribution-based explainability analysis.

\subsection{Data and Experimental Setup}
The nominal dataset is generated from a high-fidelity mission simulator providing an accurate representation of the spacecraft orbital environment and reproduces the main dynamical phenomena that affect the satellite throughout the mission.  

The simulation also accounts for variation in spacecraft parameters, including several disturbance sources acting on the attitude dynamics, such as payload disturbance torque and discrete static unbalance cases, Antenna Pointing mechanism and reaction wheel torque and solar array flexible model. Sensor imperfections are also included in the simulation, comprising gyroscope noise, delays, and scale factor, allowing the simulator to reflect operating conditions typically encountered in real missions.
The simulator operates at a base frequency of 32~Hz to capture high frequency dynamics, while the onboard computer operates at 8~Hz and includes satellite attitude (quaternions) and angular rates (radians). Data logging is performed consistently with the multi-rate architecture of the simulator, ensuring coherence between the recorded variable used for the dataset and the mission data.

The sensor readings for the gyroscope and star trackers are all given at 16~Hz. Since, the onboard time is set at 8~Hz, the CNN residual learner must perform its predictions at this rate. To match the target onboard execution rate of 8~Hz, all signals were first synchronized to a common timeline. Gyroscope and star tracker measurements were downsampled from 16~Hz to 8~Hz by stride selection, while the 32~Hz reference angular rates were interpolated directly onto the resulting 8~Hz timestamps. Sliding-window center times were then generated according to the selected window length. For each window, sensor measurements were interpolated onto the window timestamps, and the residual target was computed at the corresponding center time as the difference between the reference and gyroscope angular rates. This procedure ensured temporal consistency between the input sequences, correction targets, and intended onboard operating frequency.

\begin{figure}[tb]
     \begin{subfigure}[b]{0.56\textwidth}
         \centering
         \includegraphics[width=\textwidth]{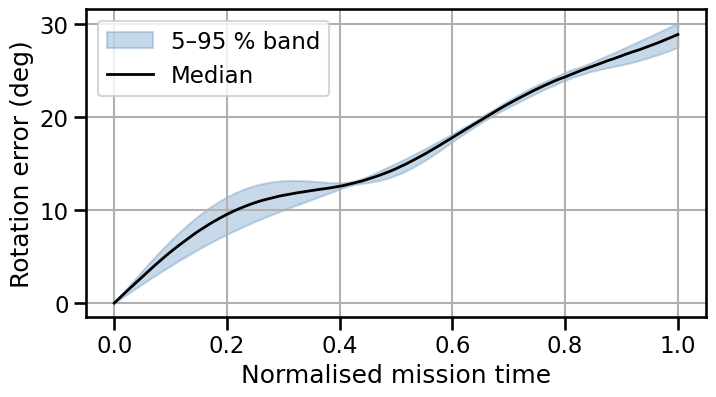}
         \caption{Gyro-only attitude drift envelope.}
         \label{fig:drift_envelope}
     \end{subfigure}
     \hfill
     \begin{subfigure}[b]{0.4\textwidth}
         \centering
         \includegraphics[width=\textwidth]{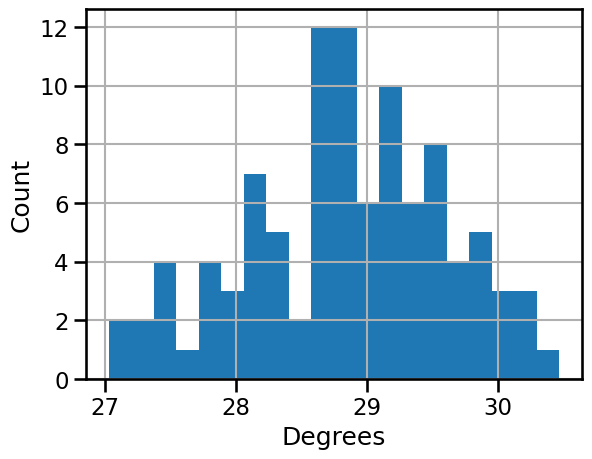}
         \caption{Final drift distribution.}
         \label{fig:final_drift}
     \end{subfigure}
   \caption{Gyro-only drift analysis across 100 logs. a) Rotation error grows monotonically, reaching approximately $30^{\circ}$ by the end of the trial. The shaded region denotes the 5--95\% percentile band. b) Final drift values remain tightly clustered between approximately $27^{\circ}$ and $31^{\circ}$.
   }
    \label{fig:gyro-only}
\vskip -0.5cm
\end{figure}

The dataset has a total of 100 logs, each with a duration of approximately $3.24$ hrs. We observe that upon a gyro-only drift analysis, the signal grows smoothly, reaching approximately $30^{\circ}$ at the end of each trial. In Figure \ref{fig:drift_envelope}, the median trajectory is shown in black and the drift envelope in blue. A final drift histogram (Fig. \ref{fig:final_drift}) shows the final gyro drift span with all the logs between $27^{\circ}-31^{\circ}$ and tightly clustered on the median of $28.8^{\circ}$. The clustering also shows the dominating constant bias instead of random noise. Without correction, even a 10 min star tracker outage causes a $\sim 1^{\circ}$ attitude error. This illustrates that the drift is roughly consistent across the 100 logs, evolving slowly and monotonically, which is well-suited for a CNN to learn.

We design and train the CNN-residual network using BOHB (Bayesian Optimization with HyperBand) \cite{Falkner2018_bohb} for hyperparameter optimization. We split the dataset into 40-30-30 percentages for training, validation, and testing. We consider ablation parameters such as windowing strategy and stride configuration.

We evaluate the model under nominal (ID) and structured perturbation regimes of increasing intensity: Low (OOD-L), Medium (OOD-M), and High (OOD-H). These regimes differ in strength and characteristics of temporally structured perturbations applied to the sensor inputs. The perturbations are designed to progressively increase in magnitude while preserving the underlying signal dynamics. 

These emulate realistic sensor degradations, including correlated noise, vibration, bias drift, and intermittent measurement loss.

\begin{table}[t]
\centering
\caption{Summary of model configuration, sequence parameters, and perturbation regimes used in the experiments.}
\label{tab:full_config}
\begin{tabular}{lc}
\toprule
\textbf{Parameter} & \textbf{Value} \\ \midrule\hline

\multicolumn{2}{|c|}{\textbf{Model and Training}} \\ \hline
CNN layers & 1--3 \\
Filters & 16--128 \\
Kernel size & 4--16 \\
Max pooling & 2--8 \\
Batch size & 16--64 \\
Dropout & 0--0.5 \\
Optimization & BOHB (validation NLL) \\
Epochs (full budget) & 150\\
Ensemble size (different seeds) &  5\\ \hline

\multicolumn{2}{|c|}{\textbf{Sequence Configuration}} 
\\ \hline
Window length & 1.0 s (8 timesteps @ 8 Hz) \\
Stride & 0.25 s (2 ts), 1.0 s (8 ts) \\
Overlap & 6 ts (overlapping), 0 ts (no overlap) \\ \hline

\multicolumn{2}{|c|}{\textbf{Perturbation Regimes}} \\ \hline
Regimes & OOD-L, OOD-M, OOD-H \\
AR noise strength ($\rho$) & 0.05 / 0.10 / 0.20 \\
Vibration scale & 0.03 / 0.06 / 0.10 \\
Frequency range (Hz) & 0.5 -- 3.8 \\
Dropout probability & $6\times10^{-4}$ -- $8\times10^{-4}$ \\
Dropout duration & 2 s / 5 s / 15 s \\
Bias step scale & 0.05 / 0.10 / 0.20 \\
Drift rate (per s) & 0.001 / 0.003 / 0.006 \\ \bottomrule

\end{tabular}
\vskip -0.5cm
\end{table}

In Table \ref{tab:full_config}, we summarize the complete configurations for training, ablation, and perturbation regime generation. Preliminary experiments without the aleatoric head explored varying temporal window lengths (0.5–1.0 s) and overlap configurations. These results indicated that a 1.0 s window provided the best trade-off between performance and stability, and was therefore used in all subsequent experiments. Therefore, for these experiments, the stride was fixed to 0.125 s (1 timestep) to isolate the effects of window size. Because the sampling frequency is constant (8 Hz), modifying the window length inherently changes the proportion of overlap. For a larger window, e.g., 1.0 s, a one-timestep stride implies very high overlap. For shorter windows, e.g., 0.5 s, the same stride produces moderate overlap. Thus, each configuration implicitly swept through different effective overlap levels. These variations allowed us to quantify the trade-off between window length, sample diversity, and the temporal redundancy induced by overlapping windows. 

\subsection{Correction Performance}

We obtain a set of high-performing models from the BOHB search, where a representative candidate is chosen, balancing correction performance and uncertainty behavior. The selected model is evaluated against the raw measurements and a naive baseline, where the mean bias is removed from the sensor readings. Residual errors, defined as the difference between ground truth and measurements, are compared using mean residual magnitude, RMSE, median, and 95th percentile. A summary of these results is presented in Table \ref{tab:correction_performance}.

\begin{table}[tb]
\centering
\vskip -0.7cm
\caption{Residual magnitude statistics (deg/s) and relative improvement.}
\label{tab:correction_performance}
\begin{tabular}{p{1.5cm}p{1.5cm}p{1.5cm}p{1.5cm}p{2.5cm}p{2.5cm}c}
\toprule
Metric & Raw & Naive & CNN & Improvement (CNN vs Raw) & Improvement (CNN vs Naive) \\  \midrule
Mean & 0.00661 & 0.00341 & 0.000622 & 90.6\% & 81.8\%\\
RMSE & 0.00733 & 0.00495 & 0.000844 &  88.5\% & 82.9\% \\
Median & 0.00591 & 0.00220 & 0.000462 & 92.2\% & 79\%\\
p95 & 0.0121 & 0.0103 & 0.00166 & 86.2\% & 83.8\%\\
\bottomrule
\end{tabular}
\end{table}

\begin{figure}[tb]
\centering
 \includegraphics[width=\textwidth]{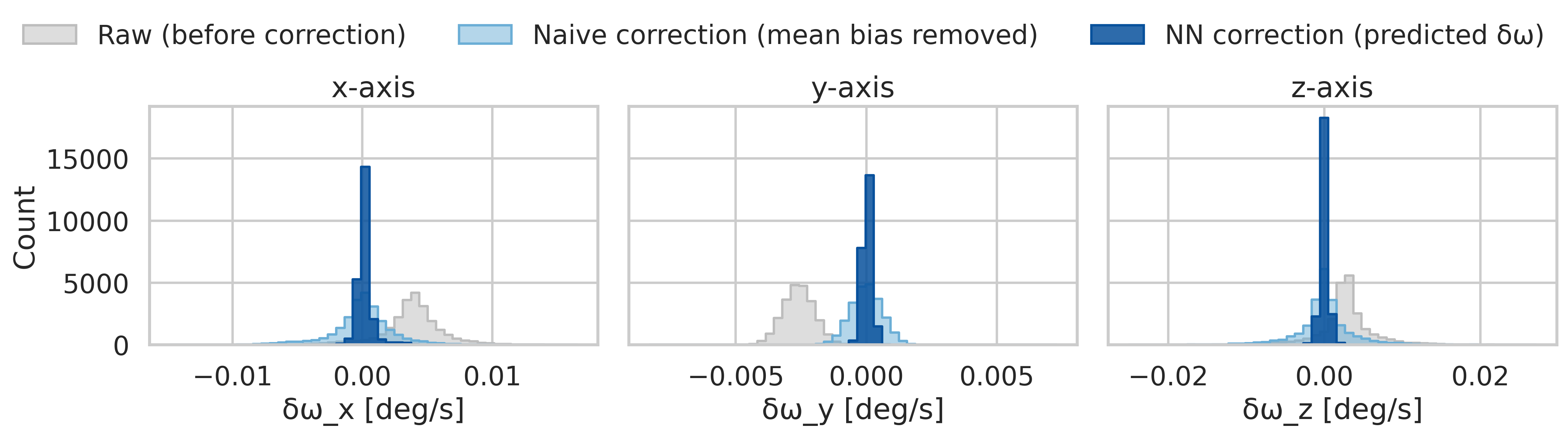}
 \caption{Residual distribution per-axis. NN-based filter correction reduces the mean magnitude of residuals by over  80\%. Plots show the residual distribution per-axis for raw measurements before correction, a naive correction approach that removes the mean bias, and the NN correction.}
\label{fig:correction_performance}
\end{figure}

Figure \ref{fig:correction_performance} shows the residual distributions for each axis. The CNN correction produces a clear contraction of the error distribution relative to both baselines. The mean residual magnitude is reduced by approximately 90\% with respect to the raw measurements and by over 80\% compared to the naive baseline. Similar reductions are observed in RMSE which indicate a consistent improvement across the full distribution. This reduction extends to higher percentiles, showing that larger deviations are also effectively corrected.

\subsection{Aleatoric Uncertainty Behavior}

We use the aleatoric outputs to build confidence intervals, $1\sigma$ and $2\sigma$ bands, and summarize average entropy for each model to evaluate that the empirical coverage matches the nominal coverage.

Under the Gaussian assumption, the first sigma band should contain 68\% of the time and the second sigma band 95\%. The selected model is closely calibrated with $1\sigma = 63.81\%$ and $2\sigma = 94.75\%$ and has a mean test entropy of $-0.5407$, which indicates relatively sharp predictive uncertainty distributions.

While these metrics indicate that the model is conservatively calibrated under nominal conditions, it does not provide insight into how uncertainty behaves during unseen OOD conditions. For this, we analyze the distribution of aleatoric uncertainty across ID and OOD regimes to assess its sensitivity to input variations.

\begin{figure}[tb]
     \begin{subfigure}[b]{0.45\textwidth}
         \centering
         \includegraphics[width=\textwidth]{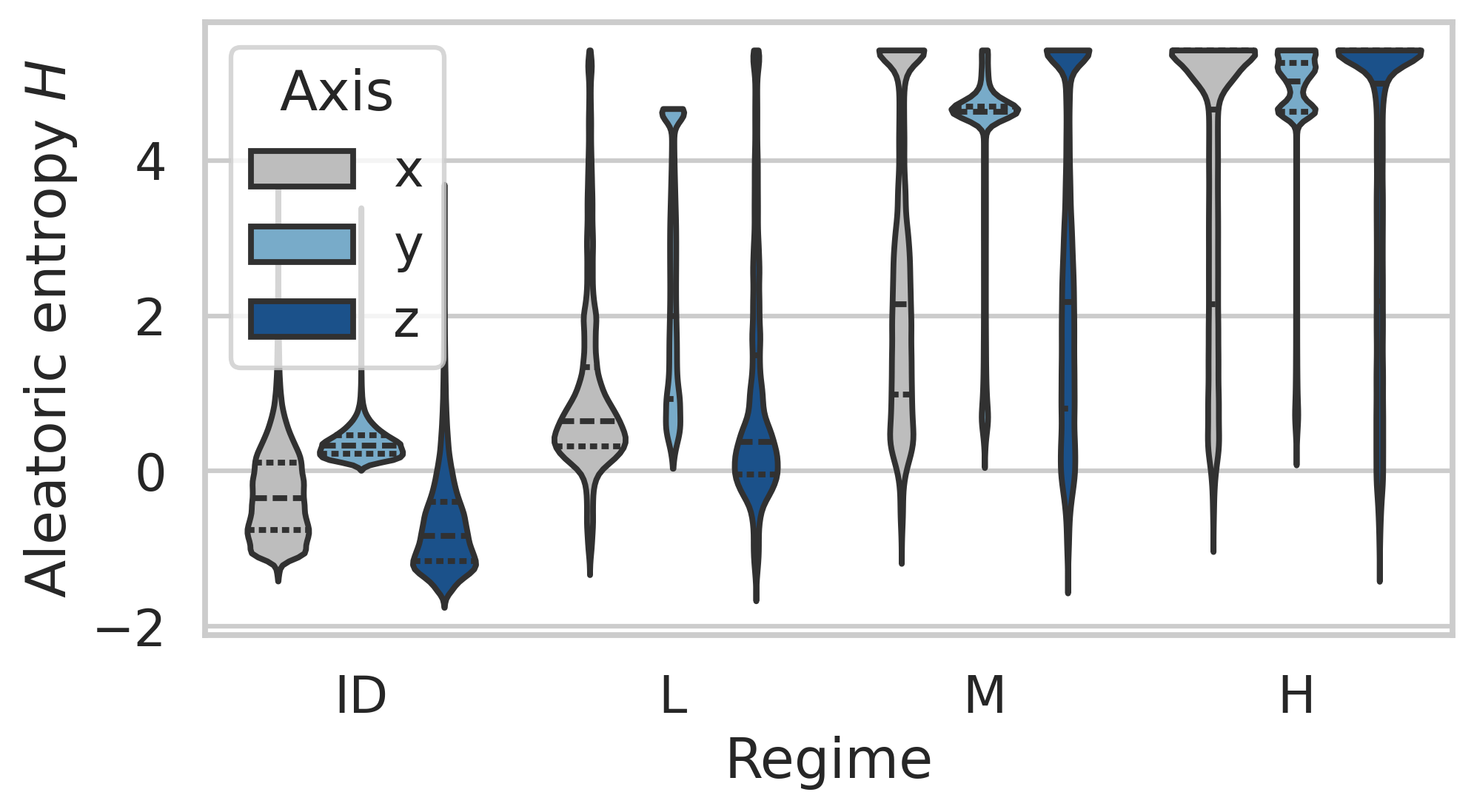}
         \caption{Aleatoric entropy distributions.}
         \label{fig:aleatoric_entropy}
     \end{subfigure}
     \hfill
     \begin{subfigure}[b]{0.53\textwidth}
         \centering
         \includegraphics[width=\textwidth]{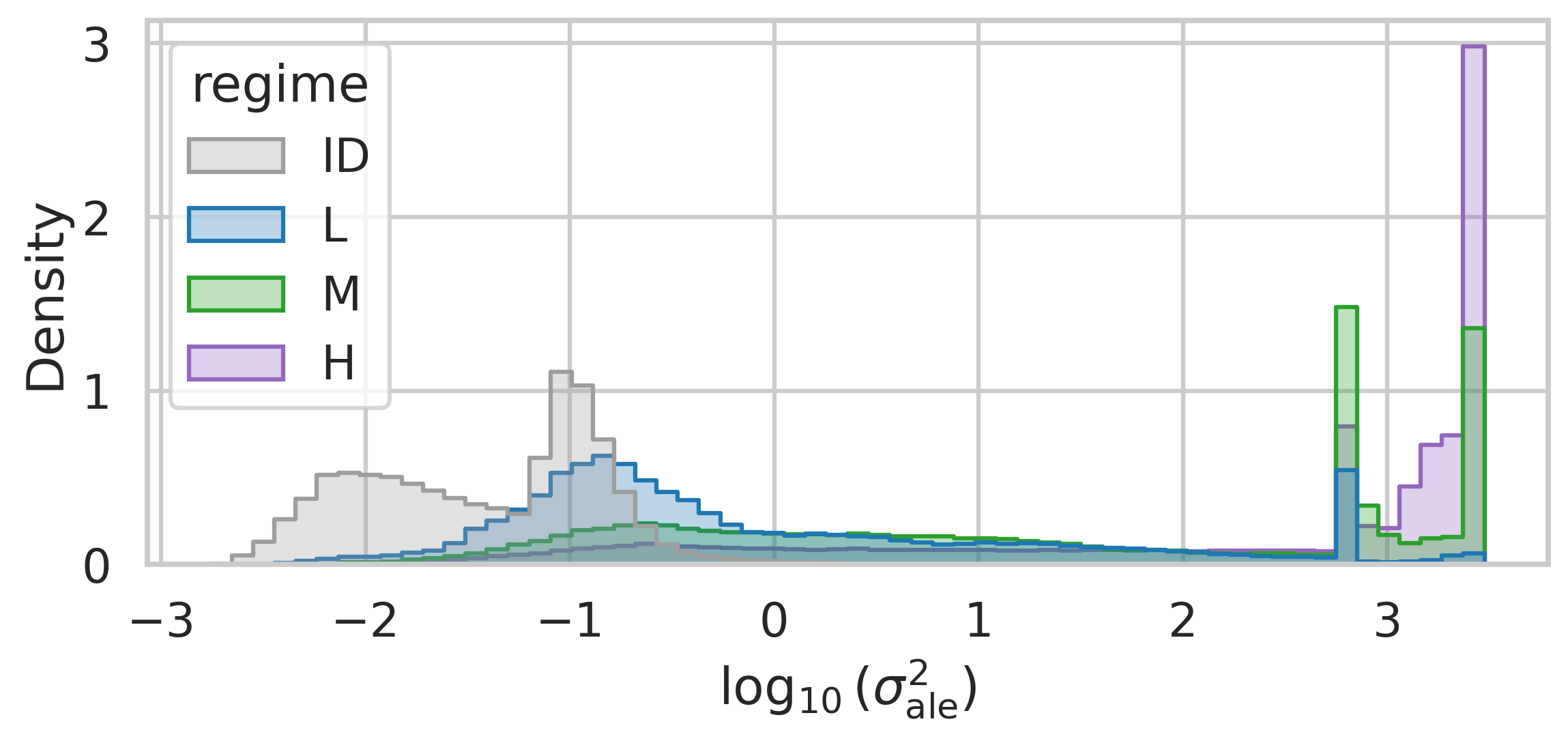}
         \caption{Predicted aleatoric variance $\log \sigma^{2}$.}
         \label{fig:logsigma_regime}
     \end{subfigure}
   \caption{Aleatoric uncertainty across regimes. a) Predicted entropy increases with perturbation intensity and exhibits broader distributions under stronger OOD conditions. b) Predicted variance shows clearer regime separation than entropy, with OOD regimes progressively shifting toward higher uncertainty values.}
    \label{fig:aleatoric_summary}
\end{figure}

\begin{figure}[tb]
     \begin{subfigure}[b]{0.4\textwidth}
         \centering
         \includegraphics[width=\textwidth]{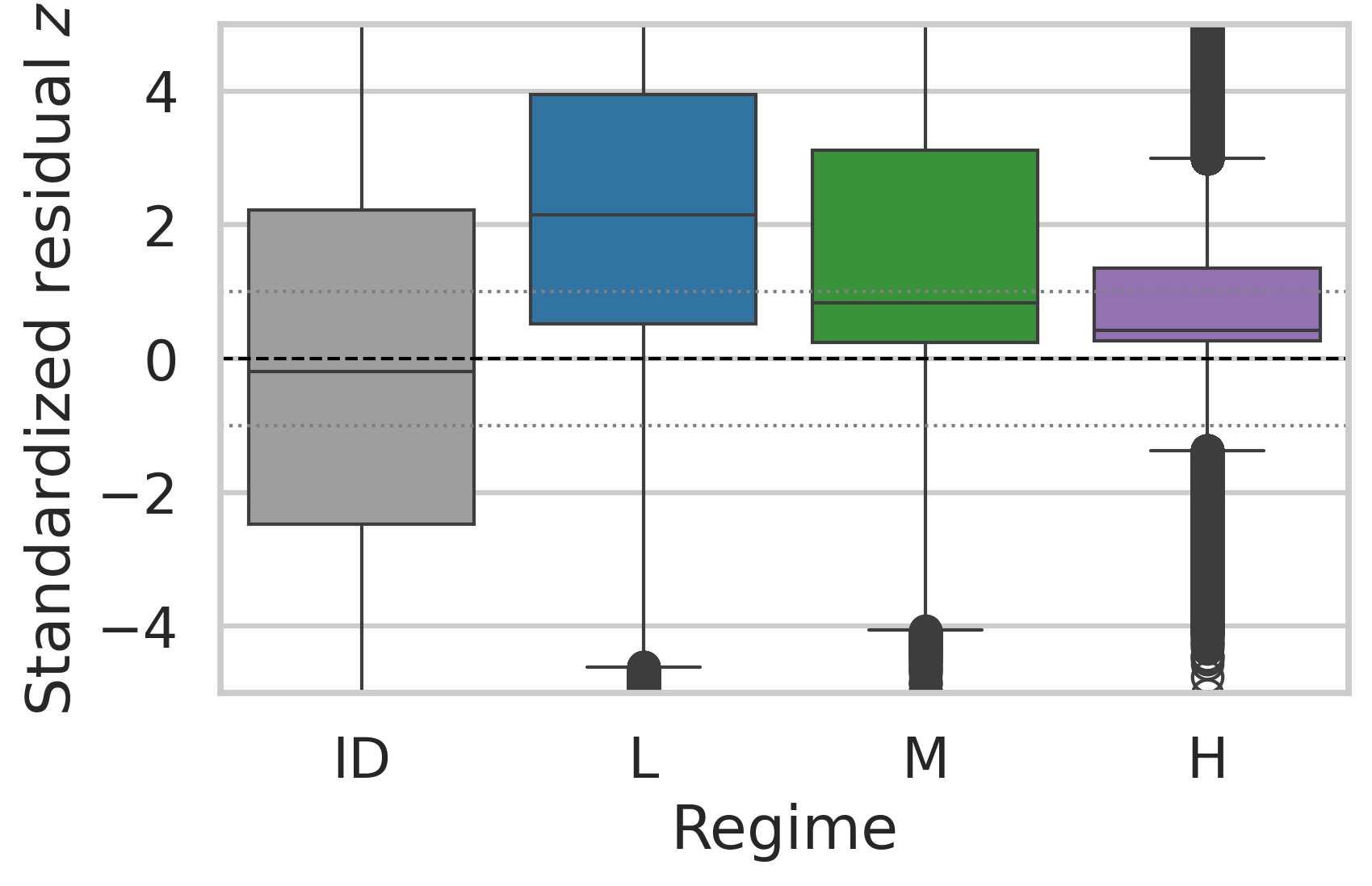}
         \caption{Distribution of $z$-scores.}
         \label{fig:boxplot_z_per_regime}
     \end{subfigure}
     \hfill
     \begin{subfigure}[b]{0.55\textwidth}
         \centering
         \includegraphics[width=\textwidth]{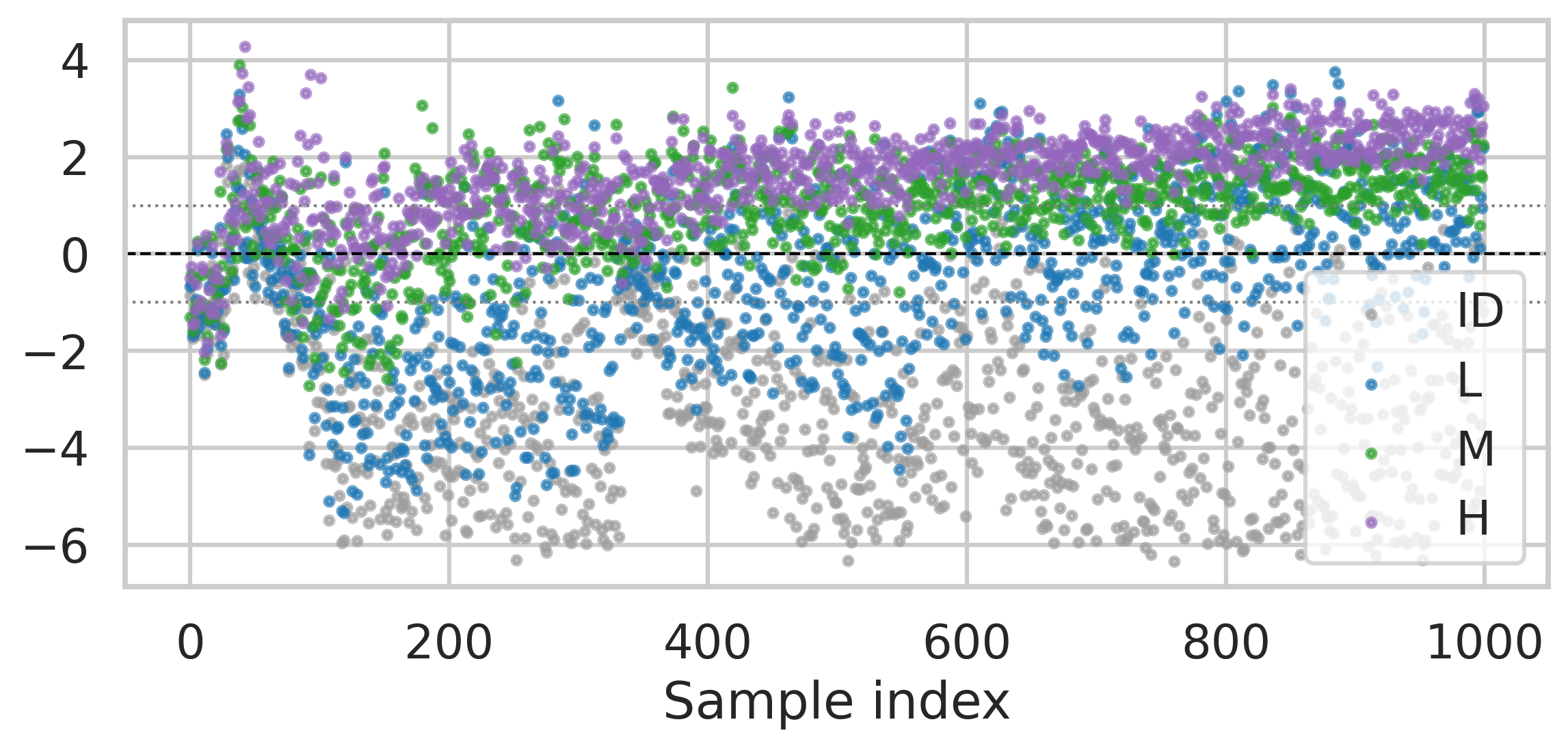}
         \caption{Sample-wise $z$-scores for all output axes.}
         \label{fig:z_timeseries}
     \end{subfigure}
   \caption{Standardized residuals ($z$-scores) distributions across regimes. 
   a) Residual distributions become progressively narrower as perturbation intensity increases, indicating increasingly conservative uncertainty estimates under OOD conditions. b) Sample-wise $z$-scores show regime-dependent shifts in residual spread and concentration.}
    \label{fig:standardized_residuals}
\end{figure}

The aleatoric uncertainty increases with perturbation intensity. In Figure \ref{fig:aleatoric_summary}, we show the shift in both entropy and variance distributions from ID to OOD regimes. This means that when the input changes in a structured way, the model's uncertainty responds by increasing.

While the distributions exhibit clear shifts, we can observe a degree of overlap remains between regimes, suggesting that the uncertainty response is not fully separable (See Figure \ref{fig:logsigma_regime}). This behavior is expected for aleatoric uncertainty, which reflects data variability rather than model confidence.

We examine the behavior of the uncertainty estimates more closely by computing standardized residuals or $z$-scores. Figure \ref{fig:standardized_residuals} shows a scatter plot and a boxplot summary for each regime. 

When conditions are normal, the residuals are spread out over a wide range, which means they are under-dispersed and the uncertainty is lower than it should be. However, the residuals get closer to zero when OOD perturbations occur. The smaller inter-quartile ranges in the boxplot and the scatter density show this. This indicates over-dispersion, signifying that the estimated uncertainty is excessively conservative. 

These results show that when there are perturbations, the uncertainty magnitude increases, but its calibration changes in a consistent way across regimes. They also show that learned uncertainty is affected by changes in input, as shown by its steady rise across regimes. However, this sensitivity does not lead to consistent calibration. Instead, the model shows under-dispersion when the inputs are nominal and over-dispersion when the inputs are OOD. This highlights the importance of assessing both uncertainty and calibration when analyzing model behavior.

\subsection{Epistemic Uncertainty Behavior}

Epistemic uncertainty is estimated using an ensemble of independently trained models. we calculate it as the variance of the predicted corrections across ensemble members, capturing model uncertainty arising from limited knowledge of the data distribution.

\begin{figure}[t]
    \centering
    \includegraphics[width=0.9\linewidth]{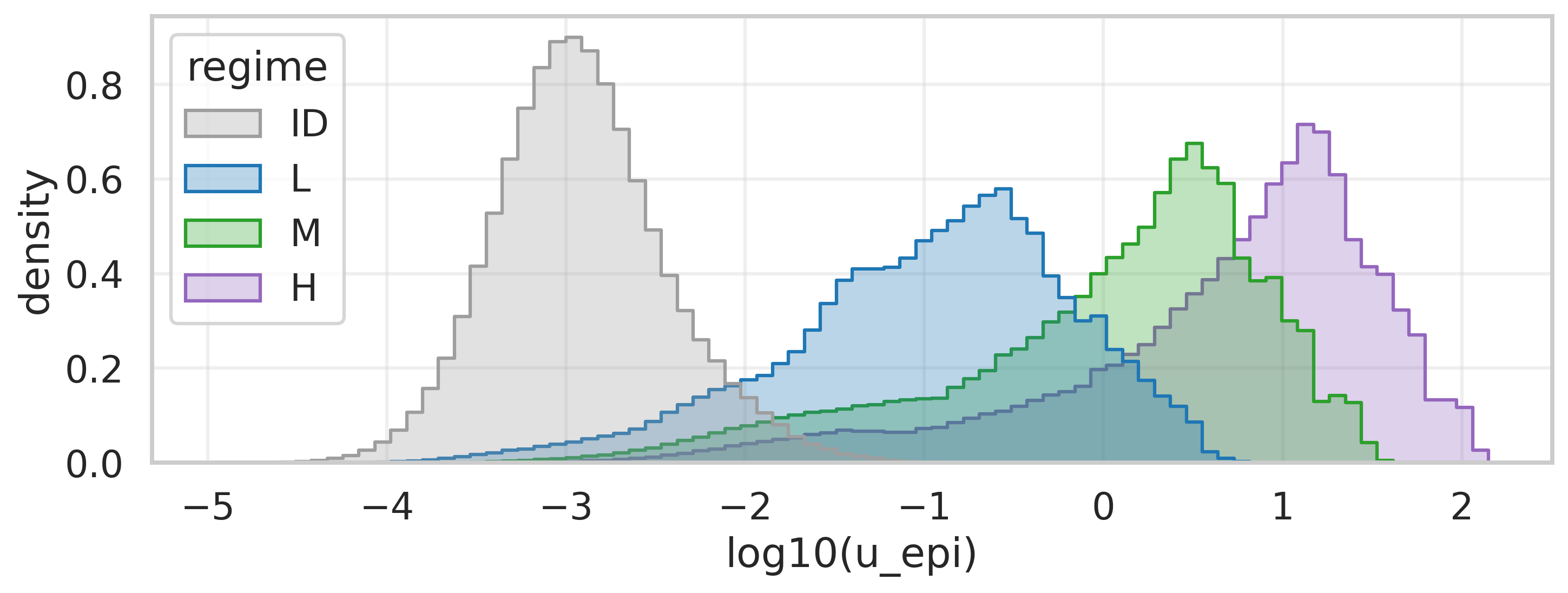}
    \caption{$\log_{10}$ epistemic uncertainty by regime. OOD distributions progressively shift toward higher uncertainty values, indicating increased ensemble disagreement under stronger perturbations.
    }
    \label{fig:epistemic}
    \vskip -0.5cm
\end{figure}

Figure \ref{fig:epistemic} shows how epistemic uncertainty is spread out across different regimes. When under nominal conditions, the ensemble variance stays low and close together. As the perturbation gets stronger, the distributions slowly move toward higher values. This makes a distinction between the nominal and OOD regimes. The overlap between regimes is very small compared to the aleatoric uncertainty. This makes it very clear how the levels of perturbation are stratified.

The distributional shift among ensemble members indicates that the model finds inputs that are not well represented in the training data. Compared to aleatoric uncertainty, these results on epistemic variance provide a direct signal of the model's confidence degradation in OOD conditions.

In contrast to aleatoric uncertainty, which showed increasing magnitude but overlapping distributions and inconsistent calibration, epistemic uncertainty provides a more reliable and separable response across regimes. This suggests that this could be more suitable for identifying non-nominal conditions in downstream monitoring and fault detection strategies.

\subsection{Attribution Analysis}

We use integrated gradients to attribute input features to both the predicted correction $\mu$ and the uncertainty output $\log \sigma^{2}$ in order to understand the learned correction and uncertainty outputs. Attributions are calculated for each axis and aggregated over time and channels to obtain interpretable feature importance and temporal patterns.

\begin{figure}[t]
     \begin{subfigure}[b]{\textwidth}
         \centering
         \includegraphics[width=\textwidth]{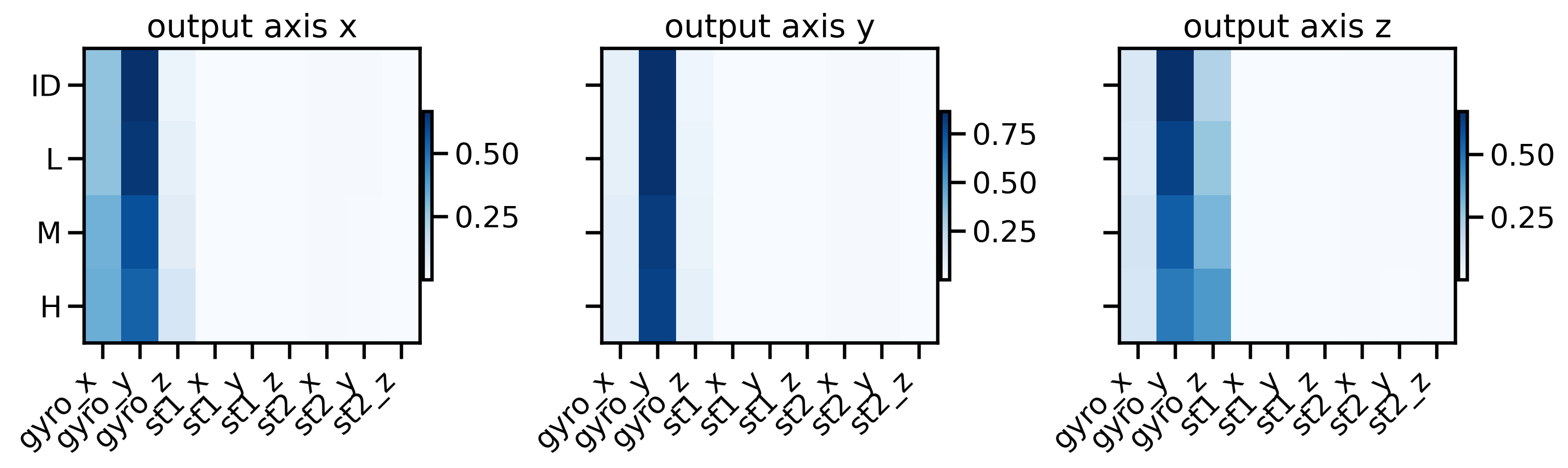}
         \caption{Channel importance for $\mu$ per regime.}
         \label{fig:channel_importance_mu}
     \end{subfigure}
     \hfill
     \begin{subfigure}[b]{\textwidth}
         \centering
         \includegraphics[width=\textwidth]{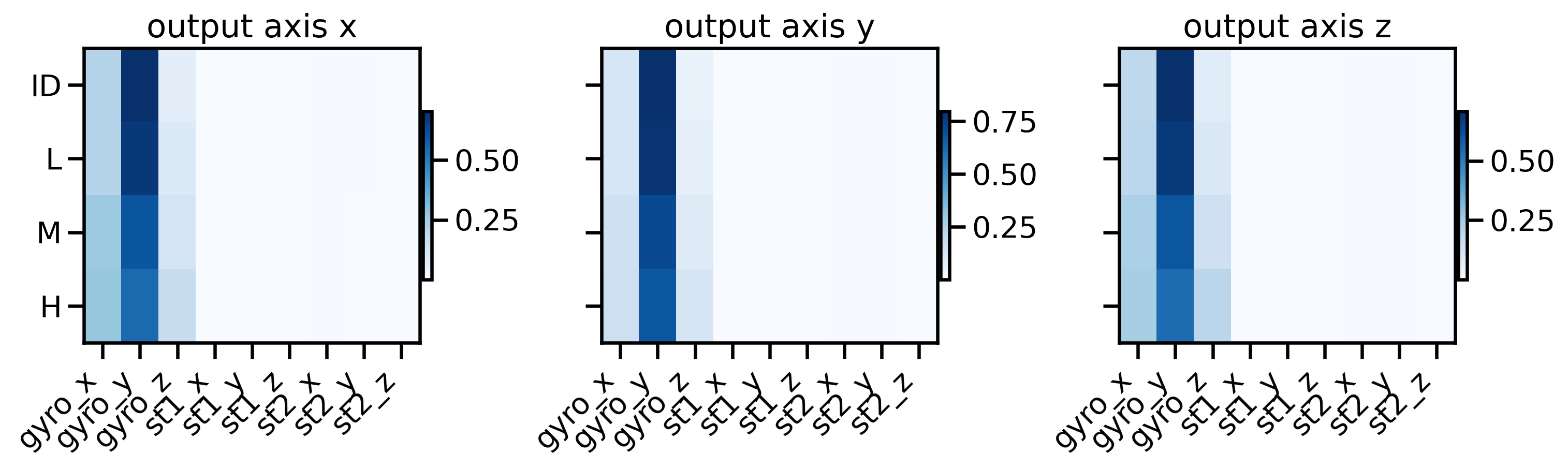}
         \caption{Channel importance for $\log \sigma^{2}$ per regime.}
         \label{fig:channel_importance_logvar}
     \end{subfigure}
   \caption{Channel attribution is dominated by gyro inputs in all regimes, particularly in the y-axis for both predicted correction $\mu$ and aleatoric uncertainty $\log \sigma^{2}$. In contrast, star tracker channels contribute negligibly.}
    \label{fig:attribution_channel}
\end{figure}

\begin{figure}[ht]
     \begin{subfigure}[b]{\textwidth}
         \centering
         \includegraphics[width=\textwidth]{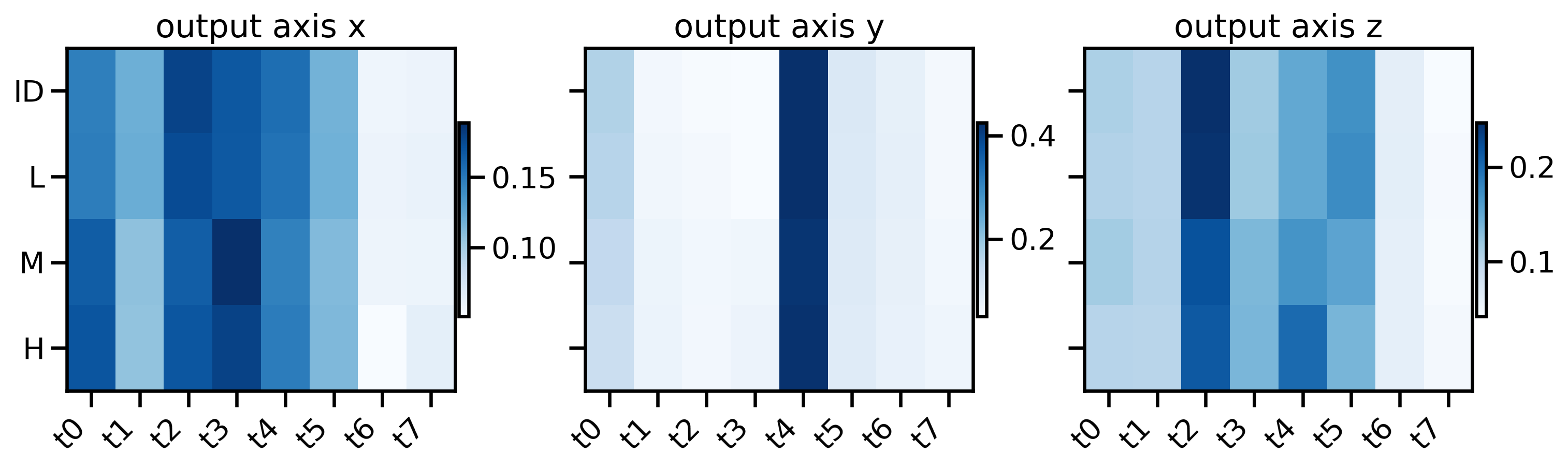}
         \caption{Temporal importance for $\mu$ per regime.}
         \label{fig:temporal_importance_mu}
     \end{subfigure}
     \hfill
     \begin{subfigure}[b]{\textwidth}
         \centering
         \includegraphics[width=\textwidth]{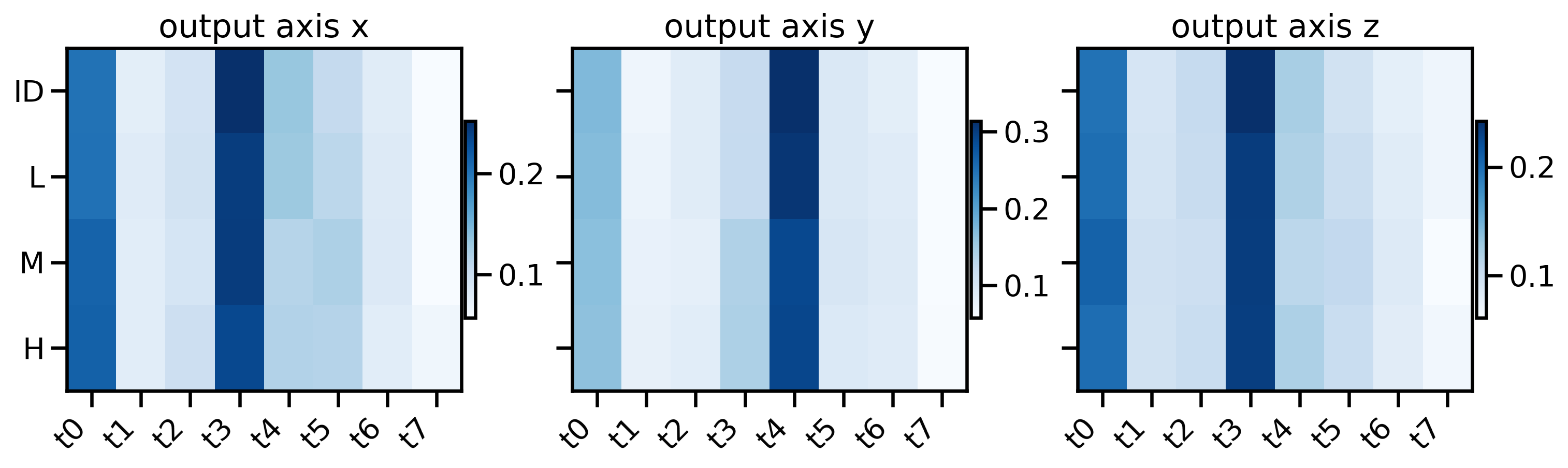}
         \caption{Temporal importance for $\log \sigma^{2}$ per regime.}
         \label{fig:temporal_importance_logvar}
     \end{subfigure}
   \caption{Temporal attribution is mainly concentrated in central timesteps, not uniformly spread. The 8-step horizon follows the selected input configuration from the BOHB ablation study and reflects the onboard 8 Hz processing rate.}
    \label{fig:attribution_time}
\end{figure}

Channel attribution reveals a strong dominance of gyroscope inputs across all regimes as shown in Figure \ref{fig:attribution_channel}. Among the gyroscope inputs, attribution is concentrated most strongly on the y-axis component across outputs in the aggregated analysis. In particular, the y-axis gyroscope consistently receives the highest attribution for both the predictive mean and aleatoric uncertainty, followed by weaker contributions from the remaining gyroscope axes. 

In contrast, star tracker channels contribute negligibly. This pattern remains stable from nominal to increasingly perturbed OOD conditions, with only moderate redistribution of importance across gyroscope axes. This indicates that the model primarily relies on inertial measurements to both estimate the correction and quantify uncertainty, with the star tracker providing an indirect attitude reference. This may be due to the fact that the model is indeed fundamentally correcting gyroscope deviations. Additionally, results suggest that uncertainty predictions are driven by the same input sources as the mean prediction.

\begin{figure}[t]
     \begin{subfigure}[b]{\textwidth}
         \centering
         \includegraphics[width=\textwidth]{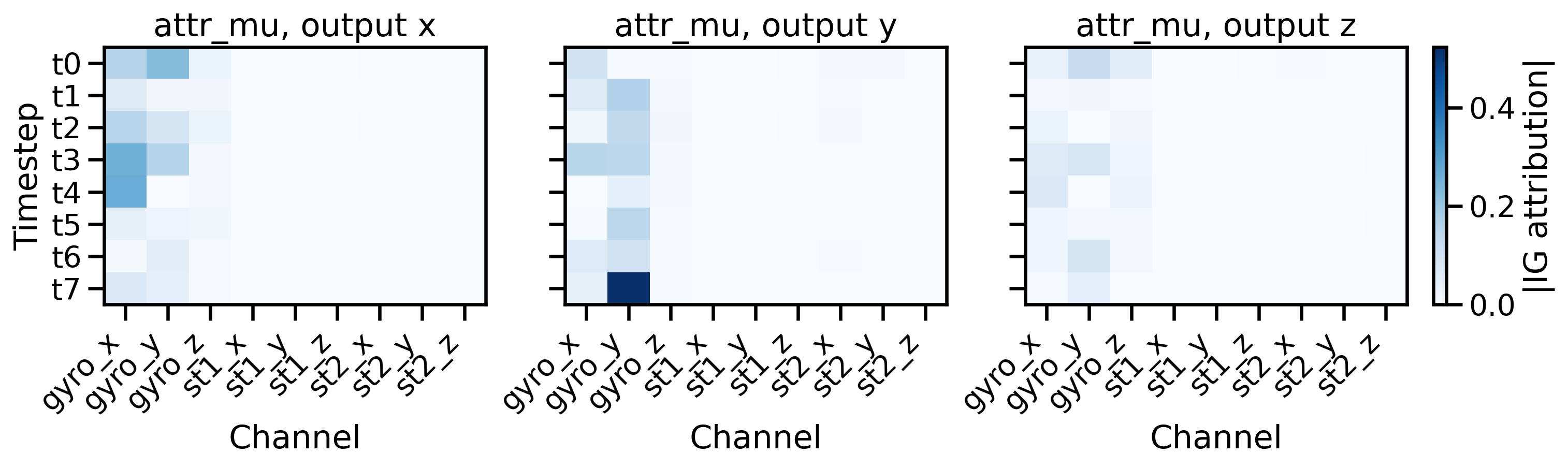}
         \caption{ID sample map for $\mu$.}
         \label{fig:debug_ID_mu_sample}
     \end{subfigure}
     \hfill
     \begin{subfigure}[b]{\textwidth}
         \centering
         \includegraphics[width=\textwidth]{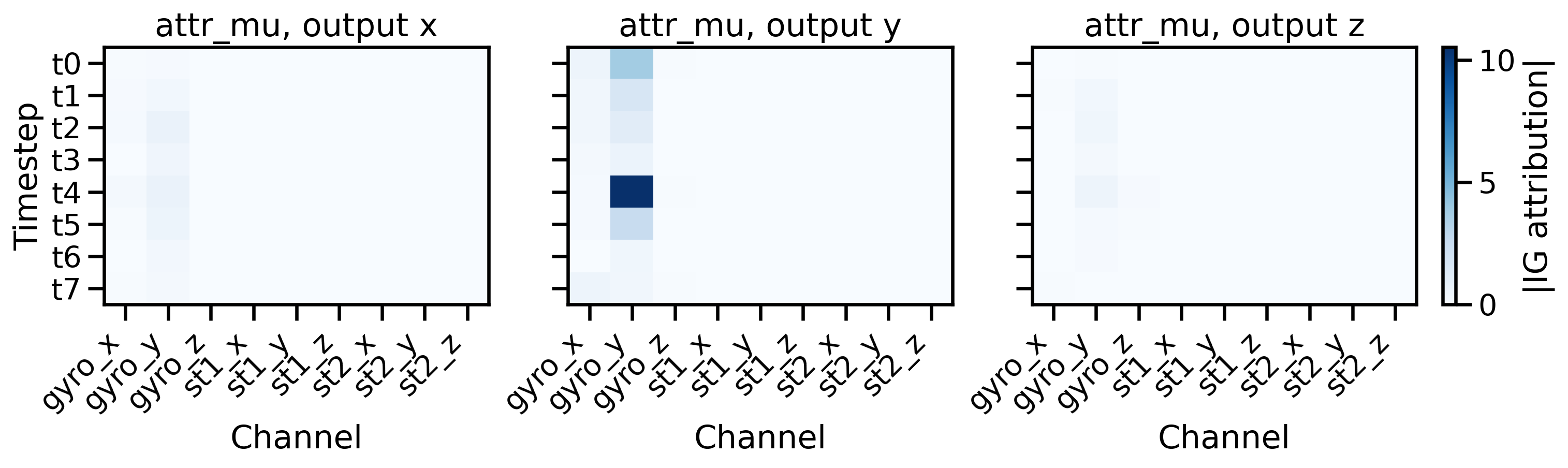}
         \caption{OOD-M sample map for $\mu$.}
         \label{fig:debug_OOD_mu_sample}
     \end{subfigure}
   \caption{Representative single-sample IG attribution maps for the predictive mean $\mu$. While the dominant attribution structure remains concentrated on gyroscope inputs, the temporal peaks and relative channel importance vary between samples and regimes, indicating local adaptation of the learned correction behavior.}
    \label{fig:single_sample}
\end{figure}

In Figure \ref{fig:attribution_time}, the temporal attribution of the predicted correction shows more axis-dependent behavior instead of a uniform pattern across outputs. For the x-axis prediction, importance is distributed more broadly across the 1s window, with contributions from both early and central timesteps. In contrast, the y-axis prediction is dominated by a single timestep around the middle of the window, indicating a highly localized dependency. The z-axis exhibits an intermediate behavior, with multiple contributing timesteps but still a clear peak. These patterns remain relatively consistent across regimes, suggesting that the model is relying on the fixed temporal structures within the input window, with limited adaptation under increasing perturbation intensity.

The temporal attribution for the aleatoric head has a structure that is similar to the mean prediction, with clear peaks at certain timesteps. However, the attribution is generally less sharply focused than $\mu$, with importance spread out over nearby timesteps instead of being focused on one main point. This effect is small and changes depending on the axis, but it shows that estimating uncertainty depends on temporal cues that are a little more spread out while still focusing on the most important parts of the input window.

Importantly, no substantial shift in temporal attribution is observed between ID and OOD regimes. Instead, the model preserves the same temporal dependencies while increasing uncertainty magnitude. This suggests that uncertainty does not arise from a change in where the model looks in time, but from increased ambiguity in the same informative regions. The model relies on the same timesteps under both nominal and perturbed conditions, suggesting that uncertainty increases due to noisier inputs rather than changes in temporal reasoning.

To complement the aggregate temporal attribution analysis, Figure \ref{fig:single_sample} shows representative single-sample attribution maps under ID and OOD conditions. These maps reveal local variation in the dominant temporal peaks and relative channel importance. However, the aggregate attribution structure remains consistently dominated by gyroscope inputs, particularly $gyro_y$.

This indicates that the network does not rely on a fixed attribution template, but instead adapts its focus depending on the local temporal structure of the input sequence. Despite this variability, the overall attribution structure remains qualitatively consistent between ID and perturbed regimes. These attribution patterns should be interpreted as model reliance rather than direct physical causality.

\section{Conclusion}

We presented an uncertainty-aware analysis of learned residual gyro correction within a hybrid state estimation framework. We designed a CNN with heteroscedastic aleatoric outputs trained under nominal data. We evaluate the network on both ID and structured OOD perturbation regimes using uncertainty calibration, ensemble disagreement, and attribution-based explainability analysis. While the OOD datasets are synthetically generated, they are structured and physically motivated but do not exhaust all possible operational distribution shift in real spacecraft sensors.

The learned correction significantly reduced residual angular-rate error relative to both raw measurements and a naive bias-removal baseline. Aleatoric uncertainty increased consistently with perturbation intensity, while epistemic uncertainty exhibited progressive separation between nominal and OOD regimes through increased ensemble disagreement. Standardized residual analysis further showed that uncertainty behavior changes systematically across operating conditions, transitioning from under-dispersed behavior under nominal conditions toward increasingly conservative uncertainty estimates under stronger perturbations.

Attribution analysis showed that the model mostly relies on gyroscope inputs for both correction and uncertainty estimation, with the $gyro_y$ component getting the most attribution across all outputs. The limited attribution assigned to star tracker channels suggests that the network primarily exploits inertial information for the residual correction task, while star tracker measurements play a secondary role.
This suggests that the CNN may be exploiting nonlinear, temporal, or cross-axis structures within the learned representation rather than simple axis-aligned dependencies. Representative single-sample attribution maps further showed local temporal variability while preserving the overall attribution structure observed in the aggregate analysis. In future uncertainty-aware GSE pipelines, these explanations could support monitoring and fault-diagnosis by providing information not only about when uncertainty increases, but also which sensor inputs are responsible for that increase.

Beyond model interpretation, the attribution analysis provides practical insight to sensor usage and uncertainty generation. These results identify which measurements contribute most strongly to both correction and uncertainty estimates, helping to reveal potential sensor redundancy, diagnose unexpected model behavior, and validate that uncertainty predictions are grounded in physically meaningful inputs. 

Overall, the results demonstrate that uncertainty and explainability analyses provide complementary insights into learned hybrid estimation behavior under distributional shift. Future work will investigate broader perturbation regimes, additional uncertainty formulations, such as direct attribution of epistemic uncertainty, and integration of uncertainty-aware monitoring within downstream state estimation and Fault Detection, Isolation and Recovery (FDIR) pipelines. Additionally, a dedicated ablation study could help determine whether the low attribution effects from the star trackers reflects redundant information or a genuine lack of dependence of those measurements. 

\subsubsection*{Acknowledgments}
\begin{sloppypar}
This work was funded by the European Aerospace Agency under the GSTP programme, activity GT1I-602SA "Artificial Intelligence techniques for spacecraft attitude control and estimation" (project acronym: AI4AOCS), contract number 4000145154/24/NL/MGu, lead by Airbus Defence and Space GmbH.
This work was partially supported by the German Federal Ministry of Research, Technology and Space (BMFTR) under the Robotics Institute Germany~(RIG).
\end{sloppypar}

\bibliographystyle{splncs04}
\bibliography{biblio}
%
% \begin{thebibliography}{8}
% \bibitem{ref_article1}
% Author, F.: Article title. Journal \textbf{2}(5), 99--110 (2016)

% \bibitem{ref_lncs1}
% Author, F., Author, S.: Title of a proceedings paper. In: Editor,
% F., Editor, S. (eds.) CONFERENCE 2016, LNCS, vol. 9999, pp. 1--13.
% Springer, Heidelberg (2016). \doi{10.10007/1234567890}

% \bibitem{ref_book1}
% Author, F., Author, S., Author, T.: Book title. 2nd edn. Publisher,
% Location (1999)

% \bibitem{ref_proc1}
% Author, A.-B.: Contribution title. In: 9th International Proceedings
% on Proceedings, pp. 1--2. Publisher, Location (2010)

% \bibitem{ref_url1}
% LNCS Homepage, \url{http://www.springer.com/lncs}, last accessed 2023/10/25
% \end{thebibliography}
\end{document}